\documentclass{bmvc2k}
\usepackage{arydshln}
\usepackage{amssymb}
\usepackage{amsmath}
\DeclareMathOperator*{\argmin}{arg\,min}
\usepackage{multirow}
\usepackage{booktabs}
\usepackage{color, colortbl}
\usepackage{xcolor}
\usepackage{pifont} 
\newcommand{\cmark}{\ding{51}}
\newcommand{\xmark}{\ding{55}}

\definecolor{lightgray}{gray}{0.9}

\title{\textit{LOCATE:} Self-supervised Object Discovery via F\underline{lo}w-guided Graph-\underline{c}ut \underline{a}nd Boo\underline{t}strapped S\underline{e}lf-training}

\addauthor{Silky Singh}{silsingh@adobe.com}{1}
\addauthor{Shripad Deshmukh}{shdeshmu@adobe.com}{1}
\addauthor{Mausoom Sarkar}{msarkar@adobe.com}{1}
\addauthor{Balaji Krishnamurthy}{kbalaji@adobe.com}{1}

\addinstitution{
Media and Data Science Research\\
Adobe
}

\runninghead{SINGH, DESHMUKH, SARKAR, KRISHNAMURTHY}{LOCATE}

\def\etal{\emph{et al}\bmvaOneDot}

\begin{document}

\maketitle
\vspace{-0.6cm}
\begin{abstract}
\label{sec:abstract}

Learning object segmentation in image and video datasets without human supervision is a challenging problem. Humans easily identify moving salient objects in videos using the gestalt principle of common fate, which suggests that what moves together belongs together. Building upon this idea, we propose a self-supervised object discovery approach that leverages motion and appearance information to produce high-quality object segmentation masks. Specifically, we redesign the traditional graph cut on images to include motion information in a linear combination with appearance information to produce edge weights. Remarkably, this step produces object segmentation masks comparable to the current state-of-the-art on multiple benchmarks. To further improve performance, we bootstrap a segmentation network trained on these preliminary masks as pseudo-ground truths to learn from its own outputs via self-training. We demonstrate the effectiveness of our approach, named \textbf{{LOCATE}}, on multiple standard video object segmentation, image saliency detection, and object segmentation benchmarks, achieving results on par with and in many cases surpassing state-of-the-art methods. We also demonstrate the transferability of our approach to novel domains through a qualitative study on in-the-wild images. Additionally, we present extensive ablation analysis to support our design choices and highlight the contribution of each component of our proposed method. The project code is available at: \href{https://github.com/silky1708/LOCATE}{https://github.com/silky1708/LOCATE}

\end{abstract}

\vspace{-0.5cm}
\section{Introduction}
\label{sec:intro}

\vspace{-0.2cm}
\begin{figure*}[h]
\begin{center}
\includegraphics[width=0.75\linewidth]{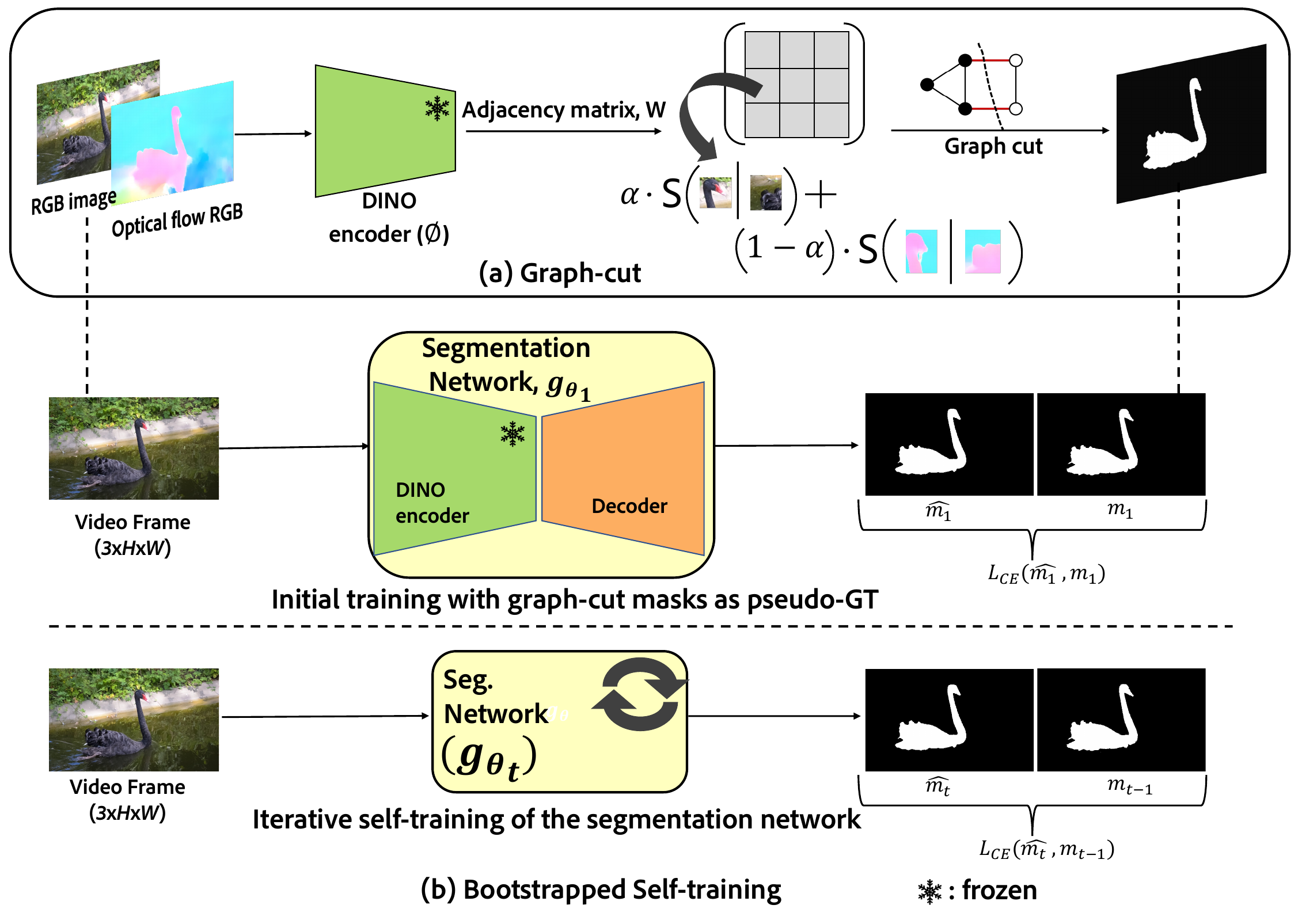}
\end{center}
\vspace{-0.5cm}
\caption{\footnotesize
\textbf{Our proposed end-to-end pipeline~(LOCATE) for self-supervised object discovery.} Our approach consists of two steps: (a) Graph-cut: we utilise a self-supervised ViT (DINO) to build a fully-connected graph on the image patches, incorporating both image and flow feature similarities to produce edge weights. We then perform graph-cut to produce a binary segmentation, followed by (b) Bootstrapped Self-training: we use graph-cut masks as pseudo-ground-truths to train a segmentation network; and subsequently train it iteratively on its own outputs.}
\label{fig:model_architecture}
\end{figure*}


Object segmentation, in its various forms, is a widely studied problem in computer vision~\cite{long2015fully}. The classic task finds critical applications across multiple domains, such as autonomous driving, augmented reality, human-computer interaction, video summarization etc. It is typically solved using deep neural networks trained on large annotated datasets created through enormous human efforts. Furthermore, dataset-specific considerations in model design limit the performance on real-world data.


To alleviate these issues with supervised segmentation, methods that work on the weaker forms of human supervision, for e.g., scribbles~\cite{lin2016scribblesup, tang2018normalized, tang2018regularized, vernaza2017learning, xu2015learning}, clicks~\cite{bearman2016s}, image-level tags~\cite{papandreou2015weakly, tang2018regularized, xu2015learning}, bounds~\cite{dai2015boxsup}, or semi-supervised segmentation techniques~\cite{dai2015boxsup, hong2015decoupled, hung2018adversarial, papandreou2015weakly, pathak2015constrained} were proposed that attempt segmentation in a setting where only a fraction of the image datasets are human-labelled. Nonetheless, both weak-supervised and semi-supervised techniques still rely on human supervision in some form.  In this work, instead of relying on supervision in any form, we aim to design an end-to-end unsupervised pipeline that can be applied to any video dataset (real or synthetic) and transfers equally well on images. This opens up critical use cases for videos in the domains such as surgery, robotics, and exploratory research in deep seas and space.


A significant chunk of object segmentation approaches tackle the problem of Video Object Segmentation~(VOS). In the literature~\cite{li2013video, ochs2013segmentation, perazzi2016benchmark, xie2022segmenting, ye2022deformable}, VOS is generally re-framed as a foreground-background separation problem, where the most salient object in the video \textit{typically} forms the foreground. Interestingly, many of these approaches are based on the principle of common fate~\cite{wertheimer2017untersuchungen}: "things that move together belong together". This notion is captured by optical flow in videos -- pixels that move with similar velocity, i.e., have similar optical flow, most likely belong to the same object. Some of the previous works~\cite{yang2021self, meunier2022driven, lamdouar2021segmenting} have advocated segmenting video objects using \textit{only} optical flow as input. Nevertheless, we hypothesize that object appearance information in addition to the motion information can provide \textit{stronger} cues for object discovery and relieve the network of ambiguities owing to occlusions or lack of motion.

So far, video and image segmentation approaches have primarily been studied disjointly. In the present work, we bridge this gap by training a segmentation network on individual frames of videos such that the resultant network can still be applied on standalone images. Our pipeline comprises of two steps. i) \textbf{Graph-cut:} For a given video frame, we build a graph(represented by adjacency matrix, $W$) with frame patches as the nodes and propose a linear combination of appearance (image) features with flow features to be used as edge weights between the nodes. We perform graph-cut on the resultant matrix $W$ to obtain a bi-partition of the image into foreground and background regions. ii) \textbf{Bootstrapped Self-training:} Next, we train a segmentation network using the acquired graph-cut masks as pseudo-ground truths. After training, the model can produce its own segmentation masks, which we then utilise for another round of training the same network from scratch. We continue this training procedure iteratively until convergence. This step allows the network to refine its predictions via self-training, and significantly boosts its ability to discover objects in videos and images. Combined together, we refer to this two-step approach by the name \textit{LOCATE} (f\textbf{LO}w-guided graph-\textbf{C}ut \textbf{A}nd boo\textbf{T}strapped s\textbf{E}lf-training). Refer Fig.~\ref{fig:model_architecture}.

Using both the steps described above, we perform a comprehensive set of experiments, achieving state-of-the-art results across multiple video and image segmentation benchmarks. We believe the simplicity of our approach paves a new path for object segmentation in videos. The contributions of this work are summarized below:
\begin{enumerate}

    \item We introduce \textbf{LOCATE}, a simple yet effective \textit{fully unsupervised} pipeline to detect salient and foreground regions in videos as well as images.

    \vspace{-0.2cm}
    \item Our approach achieves state-of-the-art results on unsupervised video object segmentation (VOS) benchmarks -- {DAVIS16} and {SegTrackv2}, while being comparable on FBMS59 dataset with state-of-the-art fully unsupervised approaches.
    
      \vspace{-0.2cm}
    \item Our segmentation network trained only on videos fares equally well on images. When evaluated in a \textit{zero-shot} manner, we are on par with or surpass current unsupervised state-of-the-art by large margins on image saliency detection (DUTS, ECSSD, OMRON) and object segmentation (CUB, Flowers-102) benchmarks.

    \vspace{-0.2cm}
    \item We provide a detailed analysis of each component of our architecture and justify the overall design choices of LOCATE.
    
    
\end{enumerate}


    
    

\vspace{-0.5cm}
\section{Related Work}
\label{sec:related_work}

\vspace{-0.15cm}
\textbf{Unsupervised Video Object Segmentation~(UVOS).} UVOS aims to localize object(s) in a video sequence by separating foreground and background regions without any annotation~\cite{tokmakov2017learning, choudhury2022guess, faktor2014video, li2018instance, lu2019see, papazoglou2013fast, tokmakov2019learning, yang2019unsupervised, yang2021dystab, xie2022segmenting, ye2022deformable}. The current state-of-the-art~\cite{choudhury2022guess} uses a quadratic component flow model to predict regions containing simple motion patterns. Since optical flow provides rich visual cues for segmenting objects in a video, many works have modelled this problem as that of motion segmentation~\cite{sundaram2010dense, keuper2017higher, keuper2015motion, keuper2018motion, tokmakov2017learning, yang2021dystab, pathak2017learning}. Layer extraction models~\cite{xie2022segmenting, ye2022deformable, chang2013topology, jojic2001learning} decompose videos into layers of persistent motion groups, but with complex inference procedures. Brox and Malik~\cite{brox2010object} cluster long-term point trajectories based on pairwise distances to achieve temporally consistent object segmentation. Faktor and Irani~\cite{faktor2014video} propose a voting scheme on co-occurring regions in the video sequence. Papazoglou and Ferrari~\cite{papazoglou2013fast} partition video frames into super-pixels and encourage foreground labels for super-pixels with independent motion. FusionSeg~\cite{dutt2017fusionseg} combines appearance and motion modalities, while MG~\cite{yang2021self} focuses on motion segmentation via slot attention mechanism. AMD~\cite{liu2021emergence} uses two visual pathways to predict coherent object regions in RGB images and optical flow vectors. Another common approach is to model optical flow as piecewise parametric motion models, such as affine or quadratic~\cite{choudhury2022guess, meunier2022driven, jepson1993mixture, torr1998geometric, mahendran2018self}.

\vspace{0.1cm}
\noindent \textbf{Unsupervised Object Discovery~(UOD).} UOD aims to discover object information from images in an unsupervised manner, for e.g., bounding boxes, object masks etc. Recent works~\cite{henaff2022object, melas2022deep, wang2022self, simeoni2021localizing, vo2021large, zhang2020object, vo2020toward} have shown the potential of using features of a pre-trained network~(supervised/self-supervised) for this task. Croitoru \etal~\cite{croitoru2019unsupervised} train a collection of student networks to mimic the performance of a teacher for unsupervised object discovery. Caron \etal~\cite{caron2021emerging} have shown that semantic information emerges in a ViT trained in a self-supervised manner. Further building on this observation, LOST~\cite{simeoni2021localizing} proposes a seed expansion strategy to localize object(s) in a given image. We aim to solve a more complex task of object segmentation, not just localization. TokenCut~\cite{wang2022self} performs spectral decomposition on dense DINO~\cite{caron2021emerging} features to segment out the most salient object in an image. Motivated by the success of this approach, we incorporate motion feature similarities with appearance features to produce edge weights.


\vspace{0.1cm}
\noindent \textbf{Object Segmentation and Image Saliency Detection.} Existing works in unsupervised/weakly-supervised image saliency detection~\cite{nguyen2019deepusps, zeng2019multi, zhang2018deep} focus on distilling hand-crafted priors into a deep network which is typically initialized with pretrained segmentation networks or image classifiers. In unsupervised object segmentation, broadly there are two categories of methods: generative or discriminative~\cite{ji2019invariant, ouali2020autoregressive} like contrastive learning. Generative approaches~\cite{benny2020onegan, arandjelovic2019object, bielski2019emergence, chen2019unsupervised} focus on generating foreground and background of an image separately that can then be combined to produce the final image. Differently, we utilise video datasets for a robust understanding of foreground objects, which can then be distilled to an image segmentation network.

\vspace{0.1cm}
\noindent \textbf{Self-training.} Self-training refers to propagating labels from a limited training set to a (large) set of unlabelled examples. This is achieved by training a network on the available labelled examples and treating its output on the unlabelled set as 'pseudo-labels' for subsequent round(s) of training. The label propagation is categorized as either hard~\cite{lee2013pseudo, xu2020iterative, yalniz2019billion} or soft assignment (often called knowledge distillation)~\cite{xie2020self} of labels. Recent works~\cite{wang2023cut, van2022discovering, zadaianchuk2022unsupervised} have shown iterative self-training to be effective while training self-supervised segmentation models on noisy pseudo-labels. In contrast, to train our network, we keep the training set fixed in consecutive rounds of training, i.e., we don't `propagate' labels to any held-out set.

\vspace{-0.4cm}
\section{Methodology}
\label{sec:methodology}



\vspace{-0.2cm}
\subsection{Preliminaries: Normalized Cut (NCut) for graph partitioning}
\label{sec:graph_cut}


\vspace{-0.2cm}
Graph-cut, in its traditional form, is a widely studied problem~\cite{shi2000normalized}. Given a graph $G = (V,E)$ where $V$ is a set of nodes and $E$ is the set of edges, it seeks to partition $G$ into two disjoint sets $P$ and $Q$. The Ncut energy as proposed in~\cite{shi2000normalized} is minimized by the optimal graph-cut, as given by:
\begin{equation}
\footnotesize
\label{eqn:1}
    \text{Ncut}(P, Q) = \frac{U(P, Q)}{U(P, V)} + \frac{U(P, Q)}{U(Q, V)}
\end{equation}

Here, $U$ is a similarity measure between two sets. $U(P, Q) = \sum_{i,j} w(p_i, q_j)$, where $p_i$ and $q_j$ are the nodes in subgraphs $P$ and $Q$ respectively and $w(p_i, q_j)$ denotes the weight between these two nodes. A diagonal matrix, $D$ can be formed with the diagonal elements given by: $d_i = \sum_{j} E_{ij}$. According to~\cite{shi2000normalized}, minimizing Eq.~(\ref{eqn:1}) is equivalent to:


\begin{equation}
\footnotesize
\label{eqn:2}
    \min_{x}~ \text{Ncut}(x) = \min_{y} \frac{y^T (D - E) y}{y^T D y}
\end{equation}
where $y \in \{1, -b\}^N$, $b$ satisfies $y^T D 1 = 0$.
Further, putting $z = D^{\frac{1}{2}}y$ in Eq.~(\ref{eqn:2}) results in the following optimization problem:
\begin{equation}
\footnotesize
\label{eqn:3}
    \min_z \frac{z^T D^{-\frac{1}{2}} (D - E) D^{-\frac{1}{2}} z}{z^T z}
\end{equation}
which is equivalent to Rayleigh quotient~\cite{golub2013matrix}. Acc. to~\cite{golub2013matrix}, the second smallest eigenvector can be used to minimize the Ncut energy in Eq.~(\ref{eqn:3}). Taking $z = D^{\frac{1}{2}}y$, the second smallest eigenvector, say $y_1$, of the following eigensystem gives a solution to the Ncut problem.

\begin{equation}
\footnotesize
\label{eqn:4}
    (D - E)y = \lambda D y
\end{equation}

To further bi-partition the graph, we use the average value of the second smallest eigenvector, $\overline{y_1} = \sum_{j} y_{1}^{j}$. Specifically, the bi-partitions $P$ and $Q$ are then given by $\{v_j ~|~ y_{1}^{j} \geq \overline{y_1} \}$ and $\{v_j ~|~ y_{1}^{j} < \overline{y_1} \}$ respectively.


\vspace{-0.1cm}
\subsection{Graph-cut for Video Object Discovery}
\label{sec:graph_cut_for_vos}

\vspace{-0.15cm}
\textbf{Graph Construction.} Consider a video frame $f \in \mathbb{R}^{H \times W \times 3}$, objects of which we wish to segment out. We create a fully-connected graph $G= (V, E)$, where $V$ is obtained by dividing $f$ into non-overlapping square patches of size $p_s\times p_s$, and $E$ denotes the edges between these patches that quantify the similarity. More specifically, an adjacency matrix $W$ underlying $E$ comprises of elements $w_{ij} = S(v_i, v_j)$, where $S(\cdot)$ is the similarity measure between two given vertices (patches).

To incorporate motion signals in the edge weights, we propose to use similarities between image patch features and corresponding optical flow(in RGB format)~\cite{liu2020learning} patch features, \textit{in a linear combination}. We obtain the `key' features from the last attention layer of a pre-trained DINO encoder~\cite{caron2021emerging}. Formally, we define $S$ as:


\begin{equation}
\footnotesize
\label{eqn:similarity}
    S(v_i, v_j) = \alpha \cdot S'(\phi(v_i), \phi(v_j)) + (1 - \alpha) \cdot S'(\phi(\psi(v_i)), \phi(\psi(v_j)))
\end{equation}

where $\alpha \in [0,1]$, $\phi(\cdot)$ denotes the DINO encoder, $\psi(\cdot)$ denotes the RGB-optical flow estimator, i.e., model computing optical flow in 3-channel RGB format, and $S'(\cdot)$ is the cosine similarity function, given by $S'(\Vec{x}, \Vec{y}) = \frac{\Vec{x}\cdot \Vec{y}}{||\Vec{x}||_2 ||\Vec{y}||_2}$. We obtain $W$ using the $S$ defined above, i.e., $W = [w_{ij}] = [S(v_i, v_j)]$.




\begin{equation}
\label{eq:6}
\footnotesize
    w_{ij} \xleftarrow{} \begin{cases}
			1, & \text{if $w_{ij}$ $\geq \tau$} \\
            \epsilon, & \text{otherwise}
		 \end{cases}
\end{equation}
Further, we normalize $w_{ij}$'s (Eq.~\ref{eq:6}) by thresholding on weight threshold hyper-parameter $\tau$. The value of $\epsilon$ is set to $10^{-5}$. Since $\epsilon \neq 0$, we obtain a fully connected graph $G$ on the patches of the frame $f$. This step is done independently for each frame in a video.

The inclusion of flow features in the graph construction differentiates our approach from other recent works~\cite{wang2022tokencut, wang2023cut}. We demonstrate that our flow-guided graph-cut approach leads to significant gains in Sec.~\ref{sec:experiments}.


\vspace{0.1cm}
\noindent
\textbf{Foreground-Background separation via Graph-cut.} We employ the Ncut algorithm~(Sec. \ref{sec:graph_cut}) to find the optimal bi-partition of image patches by solving for the second-smallest eigenvector of Eq.~(\ref{eqn:4}). To differentiate foreground patches from the background patches, we employ the heuristics proposed in~\cite{wang2023cut}: (i) the foreground patches are more dominant, hence the patch corresponding to the maximum absolute value in the second smallest eigenvector should belong to the foreground, and (ii) foreground should occupy less than two corners in the image. We then assign the value $0$ to the background patches, and $1$ to the foreground patches to get a binary object segmentation mask for frame $f$. Also, since the foreground \& background partitions are identified at \textit{patch}-level, we perform a single step of post-processing using Conditional Random Fields~(CRF)~\cite{krahenbuhl2011efficient}, to obtain binary segmentations at \textit{pixel}-level.




\vspace{-0.1cm}
\subsection{Learning to segment via Bootstrapped Self-training}
\label{sec:bootstrapped_self_training}

\vspace{-0.2cm}
Although our flow-guided graph-cut approach is able to detect foreground objects extraordinarily well (Tab.~\ref{tab:flow_model_ablation}), it still has limitations such as failing to identify small objects or to segment multiple disconnected foreground objects, etc. Thus, the segmentations obtained in Sec.~\ref{sec:graph_cut_for_vos} are {noisy}, nevertheless they still possess \textit{objectness} information that can be distilled for more robust object discovery. To that end, we propose to use these as pseudo-ground-truths to train an image segmentation network, say $g_{\theta}: \mathbb{R}^{H \times W \times 3} \rightarrow \mathbb{R}^{H \times W \times 1}$, parameterized with weights $\theta$. In our experiments, we use a pre-trained ViT (DINO) encoder with MaskFormer's~\cite{cheng2021per} segmentation head as $g_{\theta}$, but the method is applicable to all segmentation architectures. Moreover, the DINO encoder can easily be replaced with other (supervised/self-supervised) pre-trained backbones.

Given $N$ video frames, $x_{i} \in \mathbb{R}^{H \times W \times 3}$, with corresponding graph-cut masks $m_i \in \mathbb{R}^{H \times W \times 1}$, we minimze the binary cross-entropy loss $\mathcal{L}_{\text{CE}}$ in the first round of training:

\begin{equation}
\label{eq:7}
    \theta_{1}^{*} = \argmin_{\theta_1} \frac{1}{N} \sum_{i=1}^{N} \mathcal{L}_{\text{CE}} (m_i, g_{\theta}(x_i))
\end{equation}

We observe that the initial round of training improves the quality of segmentation masks, especially at the boundaries, thus boosting the segmentation performance (Sec.~\ref{exp:self_training}). After training, we infer an initial set of segmentation masks for all the images in the training set using $g_{\theta_{1}^{*}}$. Next, we use these segmentation masks as pseudo-ground-truths for training the network again from scratch. We continue this process iteratively for multiple rounds, thus bootstrapping $g_{\theta}$ by training on its own outputs as pseudo-ground-truths -- which we call \textit{Bootstrapped Self-training}. Empirically, we find that the segmentation network is able to correct its own mistakes over multiple rounds until it saturates. For subsequent rounds of training, denoted by time steps $t \in \{2,3,...\}$, we optimize the following objective function:

\begin{equation}
\label{eq:8}
    \theta_{t}^{*} = \argmin_{\theta_t} \frac{1}{N} \sum_{i=1}^{N} \mathcal{L}_{\text{CE}} (g_{\theta_{t-1}^*}(x_i), g_{\theta_t}(x_i))
\end{equation}

In recent works~\cite{wang2023cut, van2022discovering, zadaianchuk2022unsupervised}, the self-training approach has been shown effective for self-supervised training of a segmentation network with noisy pseudo-labels. These methods incorporate more data in successive rounds of training; in contrast, we train the network on a fixed dataset in each round of self-training. Thus, to avoid overfitting, we train $g_{\theta}$ from scratch rather than initializing it from previous round's checkpoint.

\vspace{-0.3cm}
\section{Experiments and Results}
\label{sec:experiments}

\vspace{-0.2cm}
\subsection{Experimental Setup}
\vspace{-0.1cm}
\textbf{Datasets.} We evaluate our approach on standard VOS benchmarks: DAVIS16~\cite{perazzi2016benchmark}, SegTrackv2 (STv2)~\cite{li2013video} and FBMS59~\cite{ochs2013segmentation}. These real-world datasets feature challenging conditions such as occlusions, motion blur, deformations, static objects etc. Following the procedure in~\cite{dutt2017fusionseg}, we merge multiple segmentation masks into one for SegTrackv2 and FBMS59 datasets for evaluation. For images, we evaluate our approach on the standard OMRON~\cite{yang2013saliency}, DUTS~\cite{wang2017learning}, ECSSD~\cite{shi2015hierarchical}, CUB~\cite{wah2011caltech} and Flowers-102~\cite{nilsback2010delving} datasets.


\vspace{0.05cm}
\noindent
\textbf{Graph-cut.} We resize the images to a resolution of 480$\times$848 for DAVIS16 and STv2 datasets, and 480$\times$640 for the FBMS59 dataset before passing through DINO, based on the dominant aspect ratio of the dataset. We use DINO's ViT-B/8 architecture for the image \& flow featurization (Sec.~\ref{exp:ablation_study}).

\noindent
\textbf{Metrics.} For comparison on the VOS task, we compute the Jaccard metric, $\mathcal{J}$ and contour accuracy, $\mathcal{F}$. $\mathcal{J}$ is the mean intersection-over-union~(mIoU) of the predicted and the ground-truth segmentation masks. For image segmentation tasks, we also compute accuracy and max${F}_{\beta}$ scores, where $F_\beta = \frac{(1 + \beta^2)~\text{precision}\cdot \text{recall}}{\beta^2 \cdot \text{precision} + \text{recall}}$. We use $\beta^2 = 0.3$. We report the sequence average of mIoU scores on DAVIS16, frame average for STv2 and FBMS59 datasets. 


\vspace{0.05cm}
\noindent
\textbf{Optical Flow estimation.} For optical flow estimation in the graph-cut step, we use an off-the-shelf model, ARFlow~\cite{liu2020learning} which is trained in a completely unsupervised fashion on the synthetic Sintel dataset. We present an ablation study in Table~\ref{tab:flow_model_ablation}.




\vspace{0.05cm}
\noindent
\textbf{Inference.} At the time of inference, our network only takes a \textit{single} image as input. Differently from existing works in VOS literature, we do not take additional inputs for e.g., optical flow, neither do we employ any test-time adaptations or post-processing techniques, rendering our approach very useful in practical applications. The runtime of our trained model is 0.42s (at an input resolution of $256 \times 512$).








\vspace{-0.35cm}
\subsection{Unsupervised Video Object Segmentation}

\begin{table}[!htbp]
\begin{center}
\footnotesize
\resizebox{\columnwidth}{!}{
\begin{tabular}{l|c |c| c| cc c c}
\toprule
\textbf{Method} & \textbf{Supervision} & \textbf{Post-Processing} & \textbf{Inference} & \multicolumn{2}{c}{\textbf{DAVIS16}~\cite{perazzi2016benchmark}} & \multicolumn{1}{c}{\textbf{STv2}~\cite{li2013video}} & \multicolumn{1}{c}{\textbf{FBMS59}~\cite{ochs2013segmentation}} \\
& & & RGB~ Flow & \multicolumn{1}{c}{$\mathcal{J}~\uparrow$} & \multicolumn{1}{c}{$\mathcal{F}~\uparrow$} & \multicolumn{1}{c}{$\mathcal{J}~\uparrow$} & \multicolumn{1}{c}{$\mathcal{J}~\uparrow$} \\
\hline
\footnotesize
SAGE~\cite{wang2017saliency} & \multicolumn{1}{l|}{\multirow{8}{*}{\parbox{2.5cm}{None\\(Fully Unsupervised)}}} & \xmark & \cmark~~~~~\cmark & \multicolumn{1}{c}{42.6} & - & \multicolumn{1}{c}{57.6} & \multicolumn{1}{c}{61.2} \\
NLC~\cite{faktor2014video} & & \xmark & \cmark~~~~~\cmark & \multicolumn{1}{c}{55.1} & - & \multicolumn{1}{c}{67.2} & \multicolumn{1}{c}{51.5} \\
CUT~\cite{keuper2015motion} && \cmark & \cmark~~~~~\cmark & \multicolumn{1}{c}{55.2} & - & \multicolumn{1}{c}{54.3} & \multicolumn{1}{c}{57.2} \\
FTS~\cite{papazoglou2013fast} && \xmark & \cmark~~~~~\cmark &  \multicolumn{1}{c}{55.8} & - & \multicolumn{1}{c}{47.8} & \multicolumn{1}{c}{47.7} \\

AMD~\cite{liu2021emergence}  & & \xmark & \cmark~~~~~\xmark & \multicolumn{1}{c}{57.8} & - & \multicolumn{1}{c}{57.0} & \multicolumn{1}{c}{47.5} \\

TokenCut~\cite{wang2022tokencut} & & CRF & \cmark~~~~~\cmark & \multicolumn{1}{c}{76.7} & - & \multicolumn{1}{c}{61.6} & \multicolumn{1}{c}{66.6} \\
Ponimatkin \etal~\cite{ponimatkin2023simple} & & CRF & \cmark~~~~~\cmark & \multicolumn{1}{c}{80.2} & 77.5 & \multicolumn{1}{c}{74.9} & \multicolumn{1}{c}{{70.0}} \\

\rowcolor{lightgray}
\textbf{LOCATE~(Ours)} && \xmark & \cmark~~~~~\xmark & \multicolumn{1}{c}{\textbf{\color{black}80.9}} & \multicolumn{1}{c}{\textbf{\color{black}88.7}} & \multicolumn{1}{c}{\textbf{\color{black}79.9}} & \multicolumn{1}{c}{68.8} \\
\midrule

MG~\cite{yang2021self}  & \multicolumn{1}{l|}{Sup. flow} & \xmark & \xmark~~~~~\cmark & \multicolumn{1}{c}{68.3} & - & \multicolumn{1}{c}{58.6} & \multicolumn{1}{c}{53.1} \\
EM~\cite{meunier2022driven} & \multicolumn{1}{l|}{Sup. flow} & \xmark & \xmark~~~~~\cmark &  \multicolumn{1}{c}{69.3} & 70.7 & \multicolumn{1}{c}{55.5} & \multicolumn{1}{c}{57.8} \\
MOD~\cite{ding2022motion} & \multicolumn{1}{l|}{Sup. flow} & DINO-based TTA & \cmark~~~~~\xmark & \multicolumn{1}{c}{79.2} & - & \multicolumn{1}{c}{69.4} & \multicolumn{1}{c}{66.9} \\
SIMO~\cite{lamdouar2021segmenting} & \multicolumn{1}{l|}{Synth. + Sup. flow} & \xmark & \xmark~~~~~\cmark & \multicolumn{1}{c}{67.8} & - & \multicolumn{1}{c}{62.0} & \multicolumn{1}{c}{-} \\
CIS~\cite{yang2019unsupervised} & \multicolumn{1}{l|}{Sup. flow} & CRF + SP & \cmark~~~~~\cmark & \multicolumn{1}{c}{71.5} & - & \multicolumn{1}{c}{62.0} & \multicolumn{1}{c}{63.5} \\
ARP~\cite{koh2017primary}  &  \multicolumn{1}{l|}{Sup. flow \& saliency} & \xmark & \cmark~~~~~\cmark & \multicolumn{1}{c}{76.3} & 71.1 & \multicolumn{1}{c}{57.2} & \multicolumn{1}{c}{59.8} \\
OCLR$^\ddagger$~\cite{xie2022segmenting} & \multicolumn{1}{l|}{Synth. + Sup. flow} & DINO-based TTA & \cmark~~~~~\cmark & \multicolumn{1}{c}{\textbf{\color{black}80.9}} & - & \multicolumn{1}{c}{72.3} & \multicolumn{1}{c}{72.7} \\
DS$^{\dagger}$~\cite{ye2022deformable} & \multicolumn{1}{l|}{Sup. flow} & \xmark & \cmark~~~~~\cmark & \multicolumn{1}{c}{79.1} & - & \multicolumn{1}{c}{72.1} & \multicolumn{1}{c}{71.8} \\
DyStaB$^{**}$~\cite{yang2021dystab}  &  \multicolumn{1}{l|}{Sup. feats.} & CRF & \cmark~~~~~\cmark & \multicolumn{1}{c}{80.0} & - & \multicolumn{1}{c}{74.2} & \multicolumn{1}{c}{{{73.2}}} \\
GWM~\cite{choudhury2022guess} &  \multicolumn{1}{l|}{Sup. flow} & CRF + SP & \cmark~~~~~\xmark & \multicolumn{1}{c}{{80.7}} & \multicolumn{1}{c}{{86.9$^*$}} & \multicolumn{1}{c}{{{78.9}}} & \multicolumn{1}{c}{\textbf{\color{black}{78.4}}} \\
\bottomrule 
\end{tabular}}
\end{center}
\vspace{-0.2cm}
\caption{\footnotesize
\textbf{Quantitative comparison with unsupervised VOS approaches.} Our method, LOCATE, achieves or surpasses the state-of-the-art on DAVIS16 and SegTrackv2 benchmarks. (Methods above the dividing line are fully unsupervised approaches. ${\dagger}$ denotes optimization per video sequence. $^{**}$DyStaB utilises supervised pre-training, ARP uses human supervision in the form of saliency maps. $\ddagger$ OCLR leverages manual annotations from large-scale Youtube-VOS data to generate synthetic data. Post-processing includes None, DINO-based test-time adaptation(TTA), CRF~\cite{krahenbuhl2011efficient} or SP(Significant Post-processing techniques like multi-step flow, multi-crop ensemble, and temporal smoothing.) In terms of supervision, these methods either use None, Synth.- supervision from synthetic data, Sup. flow indicating supervised optical flow estimation models (e.g., RAFT), or Sup. feats. indicating usage of features from a model trained in a supervised manner. Inference indicates the inputs at the time of inference- optical flow or(and) RGB image.) Best scores are shown in \textbf{bold}. $^*$ scores calculated using the open-source implementation provided by respective authors.}
\label{tab:quantitative_comparisons}
\end{table}


\begin{figure*}[]
\begin{center}
\includegraphics[width=1.0\linewidth]{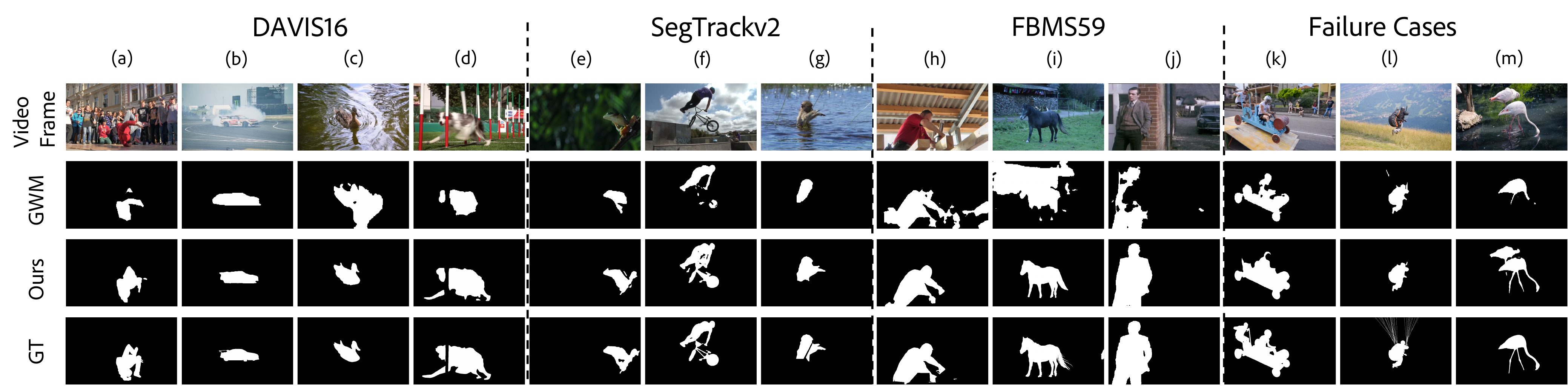}
\end{center}
\vspace{-0.6cm}
\caption{\footnotesize
\textbf{Qualitative comparison of our method on VOS benchmarks -- DAVIS16, SegTrackv2 and FBMS59.} We note that our approach is able to produce high-quality object segmentations on multiple challenging scenes. Our method is able to detect (d) occluded object(s); (a),(e),(g) deformable objects with possible camouflage in (e); (a),(h),(j) complex backgrounds; (c) doesn't get confused with reflections; (d) can handle motion blur; (f) can segment multiple foreground objects but sometimes has difficulty as shown in (k); unable to identify parachute strings in (l), and over-segmentation due to scattered attention of DINO in (m).} 
\label{fig:locate_VOS_qual}
\end{figure*}

\vspace{-0.3cm}
Table~\ref{tab:quantitative_comparisons} depicts the comparison between our method and the existing unsupervised VOS approaches. Our simple pipeline achieves state-of-the-art results on DAVIS16 and SegTrackv2 benchmarks. Compared to recent works, we also note that our approach is \textit{fully} unsupervised and does not use \textit{any} post-processing. Moreover, we do not require additional inputs like optical flow at the time of inference, thereby reducing the overhead and rendering our network applicable to image datasets as well. Despite the restricted conditions we operate in, we surpass all the recent works that utilise supervision in some form, and heavy post-processing techniques. Notably, some of the methods also utilise training on synthetic data, while we do not use supervision in \textit{any} form. On FBMS59, due to challenging scenes with multiple foreground objects, we do not perform as well, however we are still comparable to state-of-the-art fully-unsupervised methods. We show the qualitative results of our method in Fig.~\ref{fig:locate_VOS_qual}.


\vspace{-0.4cm}
\subsection{Image saliency detection and Object segmentation}
\label{exp:img_sal_obj_seg}

\vspace{-0.2cm}
\begin{figure*}[!h]
\begin{center}
\includegraphics[width=1.0\linewidth]{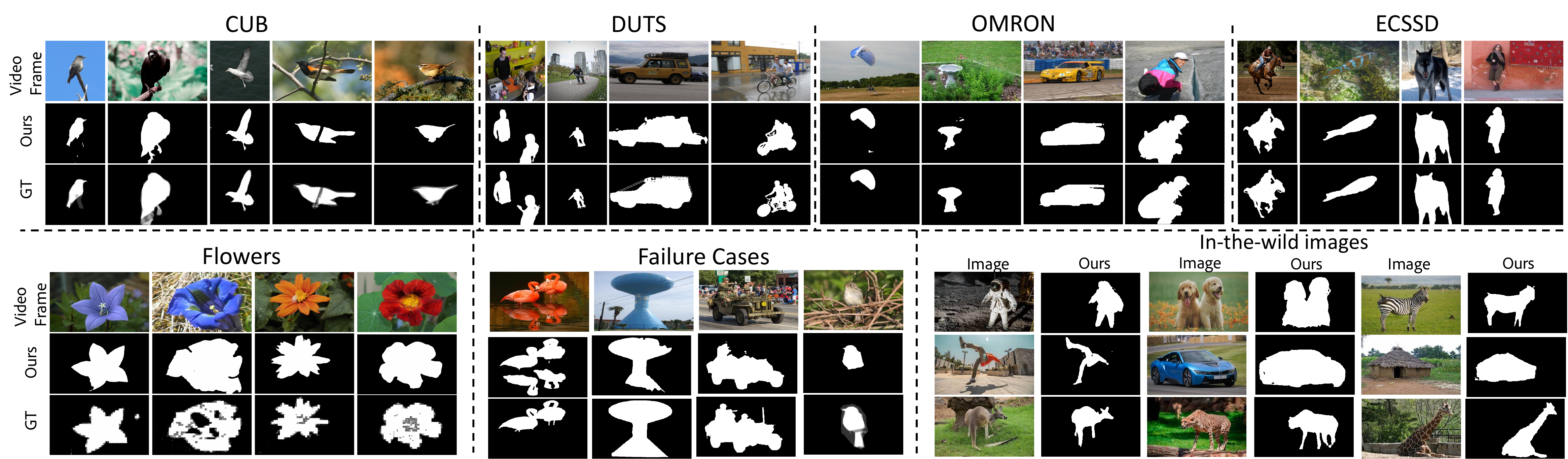}
\end{center}
\vspace{-0.6cm}
\caption{\footnotesize
\textbf{Qualitative results of our method on image saliency}~(\textbf{OMRON}~\cite{yang2013saliency}, \textbf{DUTS}~\cite{wang2017learning}, \textbf{ECSSD}~\cite{shi2015hierarchical}) \textbf{and object segmentation benchmarks}~(\textbf{CUB}~\cite{wah2011caltech}, \textbf{Flowers-102}~\cite{nilsback2010delving}). Our method applied in a \textit{zero-shot} manner is able to detect foreground objects even with complicated backgrounds, multiple foreground objects, can segment deformable objects, possible camouflaged object, flowers with fine textures and intricate boundaries. However, it sometimes gets confused with realistic reflections, fails to completely segment inanimate or occluded objects.} 
\label{fig:locate_image_qual}
\end{figure*}

\begin{table}[]
\begin{center}
\footnotesize
\resizebox{\columnwidth}{!}{
\begin{tabular}{l ccc c ccc c ccc c ccc c ccc}
\toprule
\footnotesize
 & \multicolumn{3}{c}{\textbf{CUB}~\cite{wah2011caltech}} && \multicolumn{3}{c}{\textbf{DUTS}~\cite{wang2017learning}} && \multicolumn{3}{c}{\textbf{ECSSD}~\cite{shi2015hierarchical}} && \multicolumn{3}{c}{\textbf{OMRON}~\cite{yang2013saliency}} && \multicolumn{3}{c}{\textbf{Flowers}~\cite{nilsback2010delving}} \\
 \cmidrule{2-4} \cmidrule{6-8} \cmidrule{10-12} \cmidrule{14-16} \cmidrule{18-20}
& Acc$~\uparrow$ & $\mathcal{J}~\uparrow$ & max$F_{\beta}~\uparrow$ && Acc$~\uparrow$ & $\mathcal{J}~\uparrow$ & max$F_{\beta}~\uparrow$ && Acc$~\uparrow$ & $\mathcal{J}~\uparrow$ & max$F_{\beta}~\uparrow$ && Acc$~\uparrow$ & $\mathcal{J}~\uparrow$ & max$F_{\beta}~\uparrow$ && Acc$~\uparrow$ & $\mathcal{J}~\uparrow$ & max$F_{\beta}~\uparrow$ \\
\midrule
DeepUSPS~\cite{nguyen2019deepusps} & - & - & - && 77.3 & 30.5 & 42.5 && 79.5 & 44.0 & 58.4 && 77.9 & 30.5 & 41.4 && - & - & - \\
Voynov~\etal~\cite{voynov2021object} & 94.0 & 71.0 & 80.7 && 88.1 & 51.1 & 60.0 && 90.6 & 68.4 & 79.0 && 86.0 & 46.4 & 53.3 && 76.5 & 54.0 & \textbf{76.0} \\
AMD~\cite{liu2021emergence} & - & - & - && - & - & 60.2 && - & - & - && - & - & - && -&- & - \\
Kyriazi~\etal~\cite{melas2021finding} & 92.1 & 66.4 & 78.3 && 89.3 & 52.8 & 61.4 && 91.5 & 71.3 & 80.6 && 88.3 & 50.9 & 58.3 && 79.6 & 54.1 & 72.3 \\
DeiT~\cite{touvron2021training} & - & - & - && 72.7 & 26.8 & - && 72.7 & 40.5 & - && 71.1 & 27.2 & 33.2 && - & - & - \\
Kyriazi~\etal~\cite{melas2022deep} & - & \textbf{76.9} & - && - & 51.4 & - && - & 73.3 & - && - & 56.7 & - && -& - & - \\
DyStaB~\cite{yang2021dystab} & - & -& - && - & - & - && - & - & {88.1} && - & - & \textbf{73.9} && - & - & - \\
LOST~\cite{simeoni2021localizing} & - & - & - && 88.7 & 57.2 & 69.7 && 91.6 & 72.3 & 83.7 && 81.8 & 48.9 & 57.8 && - & - & - \\
TokenCut~\cite{wang2022tokencut} & - & - & - && 91.4 & 62.4 & 75.5 && 93.4 & 77.2 & 87.4 && 89.7 & \textbf{61.8} & 69.7 && - & - & - \\
SelfMask~\cite{shin2022unsupervised} & - & - & - && 92.3 & {62.6} & - && \textbf{94.4} & \textbf{78.1} & - && 90.1 & {58.2} & - && - & - & - \\
GWM~\cite{choudhury2022guess} & 93.5 & 64.6 & 80.9 && 91.5 & 49.2 & 65.6 && 88.5 & 56.1 & 74.3 && 89.3 & 41.31 & 56.3 && - & - & - \\
DependencyViT~\cite{ding2023visual} & - & - & - && 73.2 & 35.9 & - && 78.4 & 55.0 & - && 67.2 & 28.0 & 32.5 && -& - & - \\

\midrule
\textbf{LOCATE~(Ours)} & \textbf{95.1} & {70.8} & \textbf{91.2} && \textbf{94.4} & \textbf{66.5} & \textbf{81.1} && 93.3 & 75.9 & \textbf{89.7} && \textbf{91.7} & 51.3 & {66.7} && \textbf{83.5} & \textbf{55.1} & {73.2} \\
\bottomrule 

\end{tabular}}
\end{center}
\vspace{-0.2cm}
\caption{\footnotesize
\textbf{Unsupervised image saliency detection and object segmentation.} Our model trained only on \textit{videos} compares favorably with the state-of-the-art methods when applied in a zero-shot manner on image segmentation benchmarks. Best scores are shown in bold. Scores taken from~\cite{choudhury2022guess, ding2023visual, wang2022tokencut}.}
\label{tab:img_saliency}
\end{table}

\vspace{-0.3cm}
We train our segmentation network on all \textit{three} video datasets combined- DAVIS16, STv2 and FBMS59 until convergence and evaluate in a zero-shot manner on image segmentation benchmarks. A detailed quantitative comparison with existing state-of-the-arts is provided in Table~\ref{tab:img_saliency}. Although never trained on these datasets explicitly, we are on par with or surpass all the methods. We provide qualitative results on all the benchmarks in Fig.~\ref{fig:locate_image_qual}.

\vspace{-0.3cm}
\subsection{Transferability}
\label{exp:transferability}

\textbf{In-the-wild foreground detection in images.} We conduct a transferability study of our approach on in-the-wild images. We collect a set of random images downloaded from the internet, and evaluate the pre-trained segmentation network from Sec.~\ref{exp:img_sal_obj_seg} to detect foreground object(s) in these images. The qualitative results are shown in Fig.~\ref{fig:locate_image_qual} -- under the heading ``In-the-wild images". Our model is able to segment foreground object(s) in all the challenging scenarios, demonstrating high transferability to unseen domains. Additional results are provided in the supplementary.


\vspace{0.1cm}
\noindent
\textbf{Zero-shot evaluation on unseen videos.} During iterative self-training (Sec.~\ref{sec:bootstrapped_self_training}), we train the segmentation network from scratch in each round to ensure that the model does not overfit to video frames/images of the training dataset. We evaluate the transferability of such a network by first training it on some (video) dataset, and then testing it on another (video) dataset. Empirically, we train $g_{\theta}$ on the DAVIS16 dataset, and evalute it on SegTrackv2 and FBMS59 datasets in a zero-shot manner. We find that the network is equally effective on the unseen videos of test datasets.
Further, the mean-IoU scores reported in Table~\ref{tab:zero_shot_eval} indicate that our bootstrapped self-training procedure does not deteriorate the transferability of the segmentation network.




\begin{table}[htb]
\centering
\scriptsize
\begin{tabular}{|l|c|c|}
\hline
\textbf{Datasets} & \textbf{SegTrackv2} & \textbf{FBMS59} \\ 
\hline
\hline
\textbf{mIoU~($\mathcal{J}~\uparrow$)} &  65.56 & 67.83 \\
\hline
\end{tabular}
\caption{\footnotesize
\textbf{Zero-shot evaluation on unseen videos.} Our segmentation network trained only on the DAVIS16 dataset is evaluated on the \textit{val} split of STv2 and FBMS59 datasets in a zero-shot manner.}
\label{tab:zero_shot_eval}
\end{table}

\vspace{-0.6cm}
\subsection{Trends in bootstrapped self-training}
\label{exp:self_training}

\vspace{-0.3cm}
\begin{table}[h]
\centering
\resizebox{\columnwidth}{!}{
\begin{tabular}{l|c|c|c|c|c|c|c|c|c|c|c|c}
\toprule
\multicolumn{1}{c}{} & & \multicolumn{11}{c}{\textbf{Bootstrapping Rounds}} \\
\midrule
\textbf{Datasets} & \textbf{Pseudo GT} & Round 0 & Round 1 & Round 2 & Round 3 & Round 4 & Round 5 & Round 6 & Round 7 & Round 8 & Round 9 & Round 10 \\
\midrule

\textbf{DAVIS16} & \multirow{3}{*}{\parbox{1.5cm}{Graph-cut masks}} & 78.03 & 78.98$_{\color{blue}+0.95}$ & 79.91$_{\color{blue}+0.93}$ & 80.42$_{\color{blue}+0.51}$ & 80.56$_{\color{blue}+0.14}$ & 80.61$_{\color{blue}+0.05}$ & 80.66$_{\color{blue}+0.05}$ & 80.71$_{\color{blue}+0.05}$ & \textbf{80.91}$_{\color{blue}+0.2}$ & 80.83$_{\color{red}-0.08}$ & 80.74$_{\color{red}-0.09}$ \\

\textbf{SegTrackv2} & & 75.22 & \textbf{79.94}$_{\color{blue}+4.72}$ & 79.53$_{\color{red}-0.41}$ & - & - & - & - & - & - & - & - \\

\textbf{FBMS59} & & 63.61 & \textbf{65.64}$_{\color{blue}+2.03}$ & 65.30$_{\color{red}-0.34}$ & - & - & - & - & - & - & - & - \\

\midrule
\textbf{FBMS59} & {zero-shot} & 67.83 & {68.40}$_{\color{blue}+0.57}$ & 
{68.51}$_{\color{blue}+0.11}$ & {68.75}$_{\color{blue}+0.24}$ & {68.77}$_{\color{blue}+0.02}$ & {68.8}$_{\color{blue}+0.23}$ & \textbf{68.83}$_{\color{blue}+0.03}$ & 68.74$_{\color{red}-0.09}$ & - & - & - \\ 
\bottomrule
\end{tabular}}
\vspace{0.05cm}
\caption{\footnotesize
\textbf{Bootstrapped Self-training}. We bootstrap the segmentation network by iterative self-training on its own outputs. For FBMS59 zero-shot, we start with masks obtained from a network trained on DAVIS16~(Tab.~\ref{tab:zero_shot_eval}).
}
\label{tab:bootstrapping_ablations}
\end{table}



\vspace{-0.2cm}
In Table~\ref{tab:bootstrapping_ablations}, we report the mean-IoU scores for bootstrapped self-training of segmentation networks trained on the DAVIS16, SegTrackv2 and FBMS59 datasets. The pseudo-ground-truths (scores reported under `Round 0') are either obtained through our flow-guided graph-cut approach (Sec.~\ref{sec:graph_cut_for_vos}), or in a zero-shot manner from a network trained on another dataset. We observe, in each case, that the quality of segmentation masks improves with each round of self-training until it saturates or declines in later rounds. Training on a collection of video frames enables the network to consolidate information across these examples and thus yields more accurate object segmentations.


\vspace{-0.2cm}
\subsection{Ablation Study}
\label{exp:ablation_study}

\vspace{-0.15cm}
\textbf{Encoder Model.} We test our proposed framework with three different ViT image encoders: SelfPatch~\cite{yun2022patch}, DINO-v2~\cite{oquab2023dinov2} and DINO~\cite{caron2021emerging}, and the results are reported in Tab.~\ref{tab:encoder_ablation}. We find that SelfPatch performs comparable to DINO, while DINO-v2 degrades the performance. We attribute this behaviour to the pre-training objectives of these encoders. SelfPatch primarily relies on patch-level contrastive training objectives, which leads to better patch-level representations. DINO-v2, on the other hand, loses the semantic information within an image because of its diversified pre-training procedure. Due to its superiority, we use the DINO encoder in all our experiments.

\begin{table}[!h]
\centering
\footnotesize
\begin{tabular}{|l|c|c|}
\hline
\textbf{Encoder Model} & \textbf{Graph-cut} ($\mathcal{J} \uparrow$) & \textbf{Self-training (Round 1)} ($\mathcal{J} \uparrow$)\\ 
\hline
\hline
\textbf{DINO-v2}~\cite{oquab2023dinov2} & 33.15 & \textbf{35.49} \\
\hline
\textbf{SelfPatch}~\cite{yun2022patch} & 71.64 & \textbf{73.73} \\
\hline
\textbf{DINO}~\cite{caron2021emerging} & 78.03 & \textbf{78.98} \\
\hline
\end{tabular}
\vspace{0.1cm}
\caption{\footnotesize
\textbf{Using our proposed framework with different ViT image encoders on the DAVIS16 dataset.} We use DINO-v2's ViT-B/14 and SelfPatch's available ViT-S/16 model.}
\label{tab:encoder_ablation}
\end{table}


\vspace{-0.2cm}
\noindent\textbf{DINO architecture variant.} DINO's {ViT-B} and {ViT-S} variants produce features of dimension 768 and 384 respectively. ViT-B/8, the best performing DINO architecture for graph cut (refer Table~\ref{tab:dino_ablations}), produces features of dimension 768 with an image patch size of 8.


\begin{table}[!h]
\centering
\footnotesize
\begin{tabular}{|c|c|c|c|c|}
\hline
\textbf{DINO Architecture}  & {ViT-S/8} & {ViT-S/16}   & {ViT-B/8} & {ViT-B/16}  \\ 
\hline
\textbf{DAVIS16($\mathcal{J}~\uparrow$)} & 73.46            & 74.74       & \textbf{76.76}            & 74.08                    \\
\hline
\end{tabular}
\caption{\footnotesize
\textbf{Ablation on the DINO architecture used for Graph-cut} (Refer Sec.~\ref{sec:graph_cut_for_vos}). ViT-B/8 performs the best.} 
\label{tab:dino_ablations}
\end{table}



\vspace{-0.2cm}
\noindent\textbf{Optical flow estimation model.} 
We show in Table~\ref{tab:flow_model_ablation} that our flow-guided graph-cut (Sec.~\ref{sec:graph_cut_for_vos}) is not very sensitive to the choice of optical flow estimation model. Unsupervised ARFlow~\cite{liu2020learning} performs comparable to supervised GMFlow~\cite{xu2022gmflow} on all the datasets. However to respect fully unsupervised setting, we use ARFlow in all our experiments.


\begin{table}[!h]
\centering
\footnotesize
\label{tab:ablations_flow_models}
\begin{tabular}{|l|c|c|c|} 
\hline
\textbf{Optical Flow Model} & \textbf{DAVIS16}($\mathcal{J}~\uparrow$) & \textbf{STv2}($\mathcal{J}~\uparrow$) & \textbf{FBMS59}($\mathcal{J}~\uparrow$)\\ 
\hline
\hline
\textit{GMFlow}~~\cite{xu2022gmflow} & 76.76 & \textbf{76.27} & \textbf{63.64} \\ 
\hline
\textit{ARFlow}~~\cite{liu2020learning} & \textbf{78.03} & 75.22 & {63.61} \\
\hline
\end{tabular}
\vspace{0.1cm}
\caption{\footnotesize
\textbf{Ablation on flow model used in Graph-cut.} Interestingly, we find ARFlow, which is a fully unsupervised model, performs better than its supervised alternative on DAVIS16 and fares comparable on STv2 \& FBMS59.}
\label{tab:flow_model_ablation}
\end{table}

\vspace{-0.2cm}
\noindent
\textbf{Hyper-parameters.} We find that the similarity edge threshold $\tau$ = 0.25 \& the linear combination coefficient $\alpha$ = 0.7 gives the best result (Tab.~\ref{tab:hyperparam_ablations}). Note that $\alpha = 1.0$ \& $\alpha = 0.0$ represent the cases when \textit{only} image features and \textit{only} flow features are used for graph-cut respectively. The results confirm our hypothesis that the \textit{combination} of motion and appearance information results in high-quality object segmentation masks.


\begin{table}[!h]
\centering
\footnotesize
\resizebox{0.75\columnwidth}{!}{
\begin{tabular}{|c|c|c|c|c|c|c|c|c|c|} 
\hline
\textbf{$\alpha$} & 0.0 & 0.2 & 0.4 & 0.5 & 0.6 & 0.7 & 0.8 & 0.9 & 1.0 \\
\hline

\textbf{DAVIS16}($\mathcal{J}~\uparrow$) & 46.13 & 35.22 &  56.63 & 65.20 &  72.55 & \textbf{75.96} & 75.31 & 72.19 &  62.66 \\

\hline
\hline
\textbf{$\tau$} & 0.00 & 0.05 & 0.10 & 0.15 & 0.20 & 0.25 & 0.30 & 0.35 & 0.40 \\

\hline
\textbf{DAVIS16}($\mathcal{J}~\uparrow$) & 63.05 & 71.32 & 74.38 & 75.13 & 75.96 & \textbf{76.76} & 76.56 & 75.91 & 75.83  \\ 
\hline
\end{tabular}}
\vspace{0.1cm}
\caption{\footnotesize
\textbf{Hyper-parameter ablations.} Ablation on edge threshold $\tau$ \&  linear combination coefficient $\alpha$, Sec.~\ref{sec:graph_cut_for_vos}.}
\label{tab:hyperparam_ablations}
\end{table}

\vspace{-0.6cm}
\section{Conclusion}
\label{sec:conclusion}

\vspace{-0.2cm}
We present a comprehensive study of unsupervised object discovery in videos and images via graph-cut and bootstrapped self-training. Our simple approach outperforms strong baselines on multiple standard video and image benchmarks. Moreover, we qualitatively show that our approach is equally effective in the wild. Finally, we re-emphasize that motion information present in videos can be leveraged for more robust object discovery.



\clearpage
\newpage
\vspace{0.3cm}
\section{Supplementary}
\label{sec:appendix}


\vspace{-0.2cm}
\subsection{Experimental Setup}

\textbf{Network architecture.} Similar to GWM~\cite{choudhury2022guess}, we modify the \textit{PixelDecoder} in MaskFormer's segmentation head by appending the layers \textit{[Conv(3), UpsampleNN(2), Conv(1)] $\times$ 2} to the output layer to get the output segmentation mask at the same resolution as the input. Also, since we directly obtain object segmentations through the network, we set the number of object queries to $1$, which results in a single-channel output. Further, we take \textit{sigmoid}$(x) = \frac{1}{1 + e^{-x}}$ on the output of the network ($g_{\theta}$) to produce values in the range $[0,1]$. We use a threshold of $0.5$ in all our experiments to produce a binary segmentation mask.

\vspace{0.2cm}
\noindent
\textbf{Training Setup.} All the images are interpolated to a resolution of $256 \times 512$ (using bi-cubic interpolation), before passing to the segmentation network while training. At the time of loss computation, we also interpolate the pseudo-ground-truths to $256 \times 512$ (using nearest interpolation). We employ the binary cross entropy loss function to optimize the weights of the segmentation network, $g_{\theta}$.
We use AdamW~\cite{loshchilov2017decoupled} optimizer with a base learning rate of $1.5e-4$, linearly decaying at a rate of $0.01$ starting from $1e-6$ for $1.5k$ iterations. Moreover, we train the network until convergence. Empirically, we found $25k$ iterations to be sufficient. We use a single $80$GB $A100$ GPU for training the network with a batch size of $8$.

\vspace{0.2cm}
\noindent
\textbf{Optical Flow computation in graph-cut.} Let's denote the frames of a given video by the sequence, $f_1, f_2, ..., f_N$. For a frame $f_i$, we compute the optical flow between $f_i$ and $f_{i+1}$ for $1 \leq i < N$. For $i=N$, we take the optical flow between $f_N$ and $f_{N-1}$ in our graph-cut step. The obtained optical flow is a 2-channel tensor indicating displacement of pixels in horizontal and vertical directions. We convert these to 3-channel tensors (in RGB format) using open-source implementations, for e.g., 
\href{https://github.com/ChristophReich1996/Optical-Flow-Visualization-PyTorch}{https://github.com/ChristophReich1996/Optical-Flow-Visualization-PyTorch}.


\newpage
\subsection{Qualitative Results}

\vspace{1cm}

\begin{figure*}[!h]
\begin{center}
\includegraphics[width=1.0\linewidth]{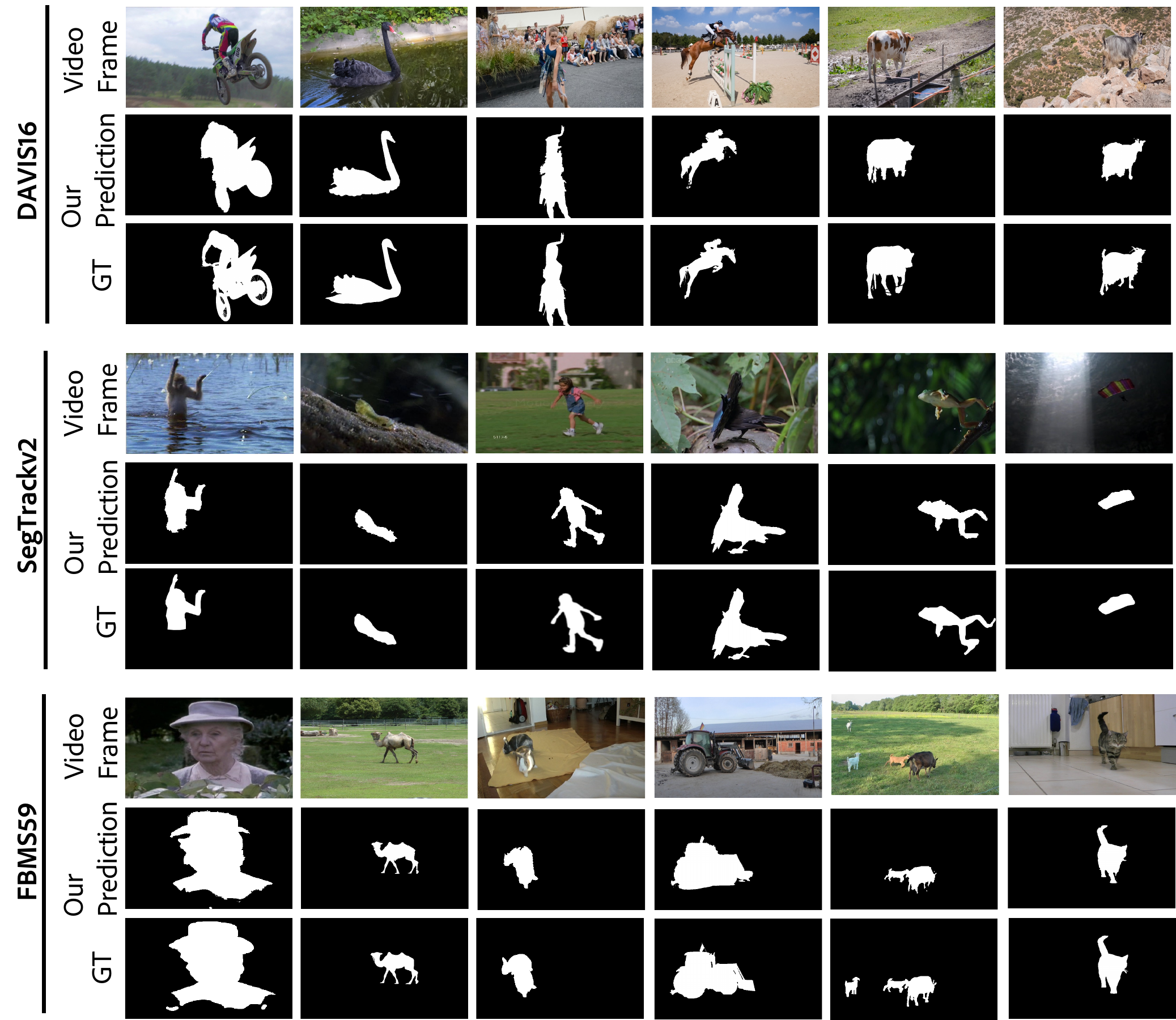}
\end{center}
\caption{\textbf{Qualitative results of our flow-guided graph-cut approach on all the video benchmarks - DAVIS16~\cite{perazzi2016benchmark}, SegTrackv2~\cite{li2013video} and FBMS59~\cite{ochs2013segmentation}.} Our approach incorporating motion information in traditional graph-cut produces high quality object segmentation masks. Quantitatively, this step alone produces results comparable to current state-of-the-art methods on DAVIS16 and STv2 datasets.}
\label{fig:graphcut_all_qual_res}
\end{figure*}

\newpage
\vfill

\begin{figure*}[!h]
\begin{center}
\includegraphics[width=0.8\linewidth]{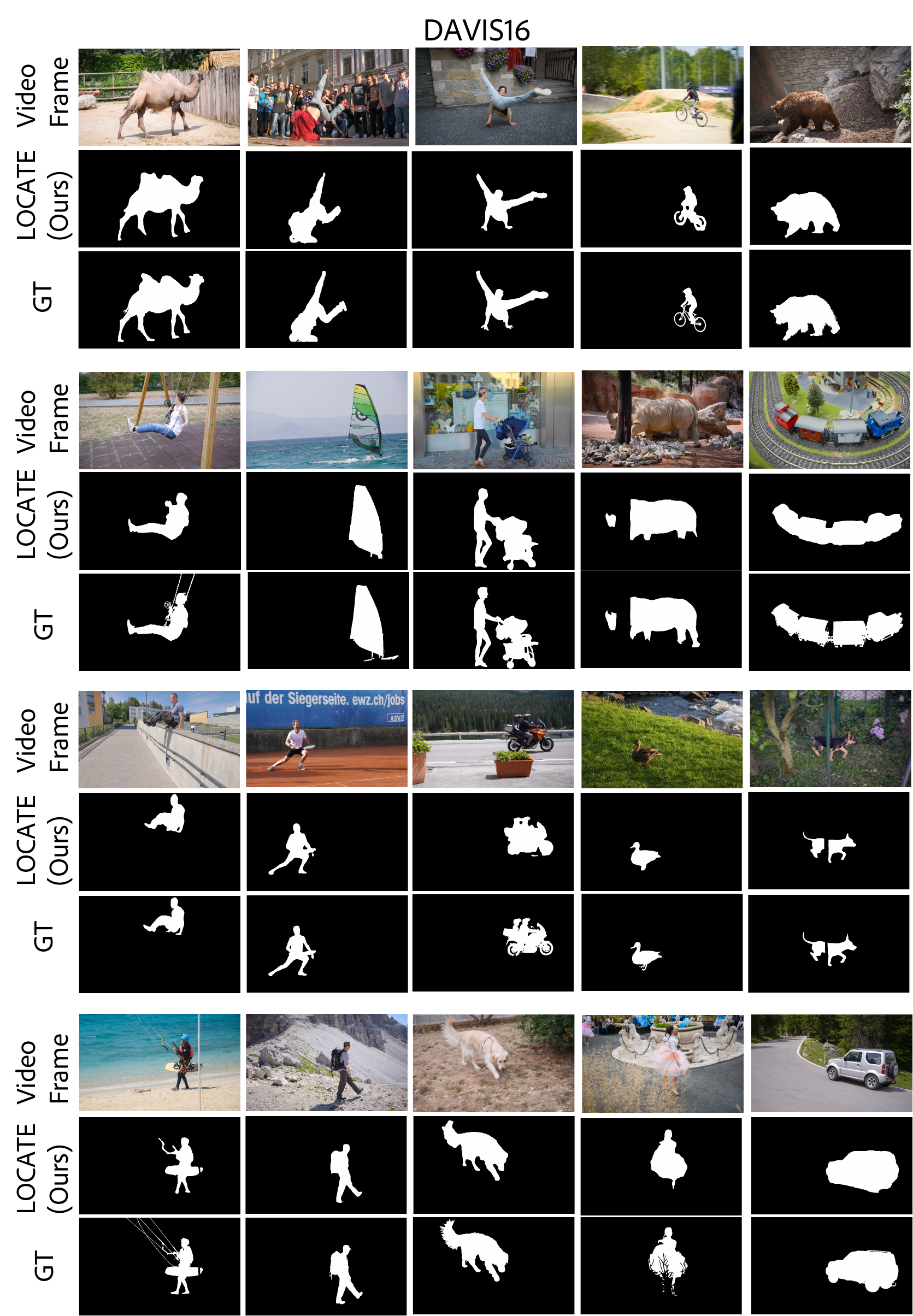}
\end{center}
\caption{\textbf{Qualitative results of our full method (LOCATE) on DAVIS16~\cite{perazzi2016benchmark} benchmark.}}
\label{fig:davis_qual_res}
\end{figure*}

\newpage
\vfill

\begin{figure*}[!h]
\begin{center}
\includegraphics[width=0.8\linewidth]{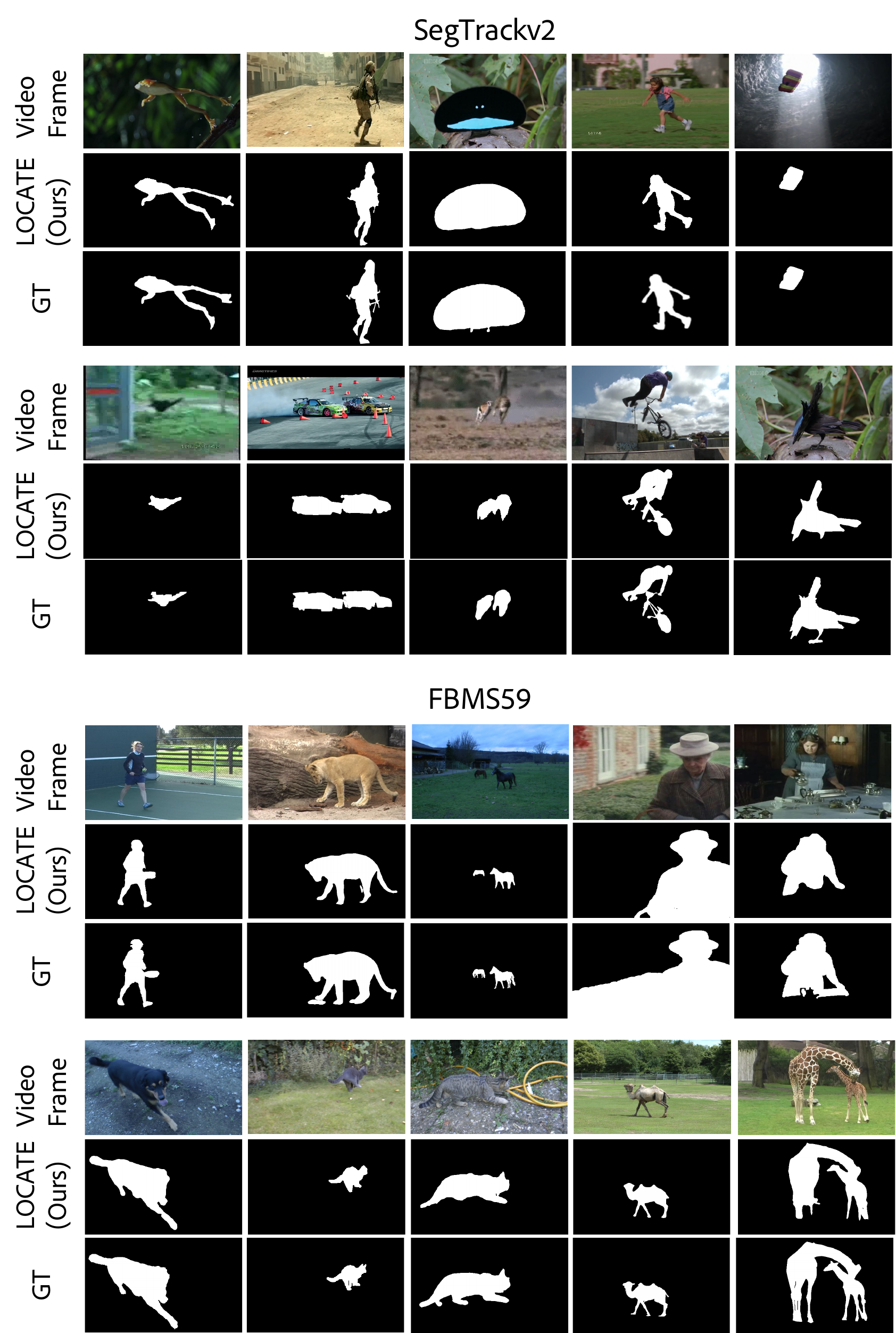}
\end{center}
\caption{\textbf{Qualitative results of our full method (LOCATE) on SegTrackv2~\cite{li2013video} and FBMS59~\cite{ochs2013segmentation} datasets.}}
\label{fig:fbms_stv2_qual_res}
\end{figure*}

\newpage

\begin{figure*}[!h]
\begin{center}
\includegraphics[width=0.8\linewidth]{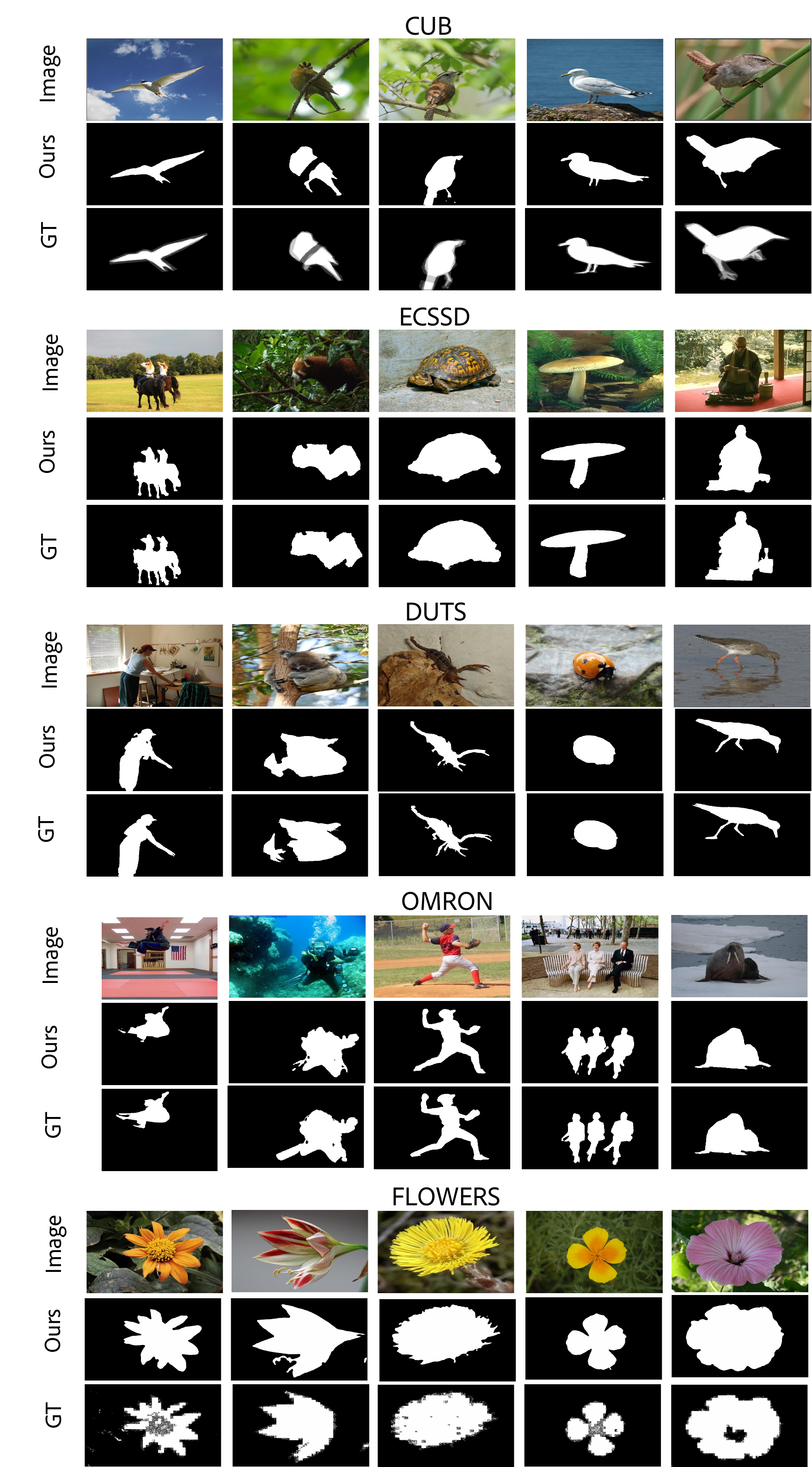}
\end{center}
\caption{\textbf{Qualitative results of our method on image saliency detection (ECSSD~\cite{shi2015hierarchical}, DUTS~\cite{wang2017learning}, OMRON~\cite{yang2013saliency}) and object segmentation (CUB~\cite{wah2011caltech}, Flowers-102~\cite{nilsback2010delving}) benchmarks.}}
\label{fig:qual_img_sal_obj_seg}
\end{figure*}

\newpage

\begin{figure*}[!h]
\begin{center}
\includegraphics[width=1.0\linewidth]{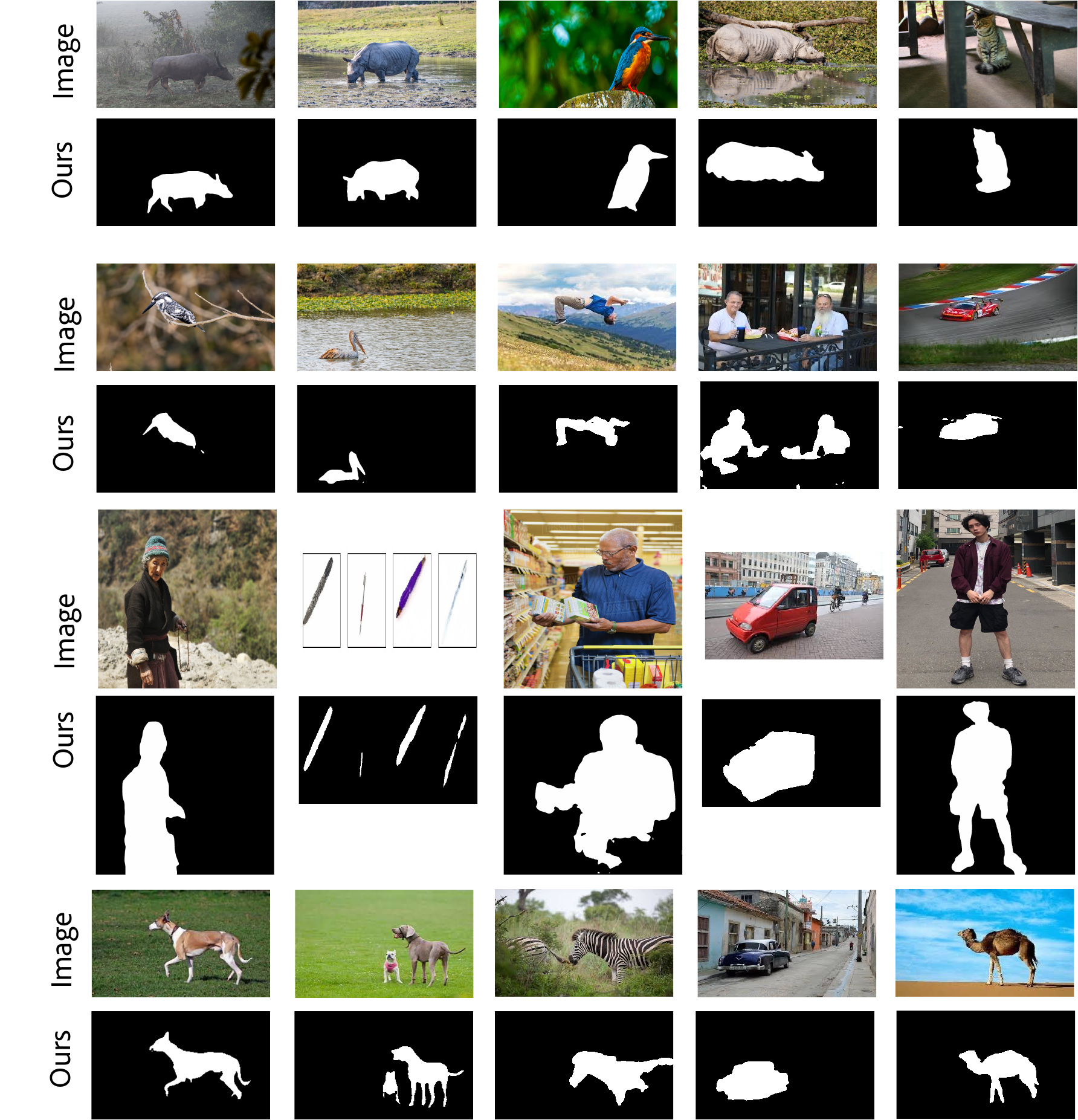}
\end{center}
\caption{\textbf{Qualitative results of our method (LOCATE) on in-the-wild images.} We asked several users to test our model on random images of their own preference, and collected the results. We show some of the representative examples and their corresponding predicted segmentation masks above. This study reinforces the effectiveness of our model in the wild.}
\label{fig:in_the_wild}
\end{figure*}
\clearpage
\bibliography{egbib}


\end{document}